# Dynamic Multi-Domain Knowledge Networks for Chest X-ray Report Generation

Weihua Liu, Youyuan Xue, Chaochao Lin, Said Boumaraf et al.

AthenaEyesCO.,LTD. & Beijing Institute of Technology

## Abstract

The automated generation of radiology diagnostic reports helps radiologists make timely and accurate diagnostic decisions while also enhancing clinical diagnostic efficiency. However, the significant imbalance in the distribution of data between normal and abnormal samples (including visual and textual biases) poses significant challenges for a data-driven task like automatically generating diagnostic radiology reports. Therefore, we propose a **D**ynamic **M**ulti-**D**omain **K**nowledge(DMDK) network for radiology diagnostic report generation. The DMDK network consists of four modules: Chest Feature Extractor(CFE)， Dynamic Knowledge Extractor(DKE)，Specific Knowledge Extractor(SKE), and Multi-knowledge Integrator(MKI) module. Specifically, the CFE module is primarily responsible for extracting the unprocessed visual medical features of the images. The DKE module is responsible for extracting dynamic disease topic labels from the retrieved radiology diagnostic reports. We then fuse the dynamic disease topic labels with the original visual features of the images to highlight the abnormal regions in the original visual features to alleviate the visual data bias problem. The SKE module expands upon the conventional static knowledge graph to mitigate textual data biases and amplify the interpretability capabilities of the model via domain-specific dynamic knowledge graphs. The MKI distills all the knowledge and generates the final diagnostic radiology report. We performed extensive experiments on two widely used datasets, IU X-Ray and MIMIC-CXR. The experimental results demonstrate the effectiveness of our method, with all evaluation metrics outperforming previous state-of-the-art models.

# 1. Introduction

Today, medical imaging plays an increasingly important role in clinical practice. On one hand, medical imaging provides physicians with substantial physiological information, including details about organs, tissues, and cells, among others, to gain an in-depth understanding of a patient's physical state. On the other hand, medical imaging can aid physicians in identifying early symptoms and irregular disease changes, thereby enhancing disease diagnostic accuracy. However, writing a proficient diagnostic radiology report requires not only extensive clinical knowledge and medical experience but also a significant amount of time and effort. Automatic generation of diagnostic radiology reports improves diagnostic efficiency and avoids inconsistencies in disease diagnosis effectively. A full medical report typically comprises several lengthy diagnostic sentences, illustrated in Fig. 1. Therefore, a functional generative report model necessitates the following critical features: (1) a report that flows naturally to match human reading habits and (2) accurate clinical diagnosis to correctly identify diseases and their related symptoms. In recent years, the progress in deep learning and natural language processing techniques has led to a surge in researchers proposing various data-driven neural networks for generating medical imaging diagnostic reports, achieving significant success. The research indicates that pertinent data can be efficiently retrieved from medical images and converted into comprehensible diagnostic reports with the aid of deep learning and natural language processing techniques.

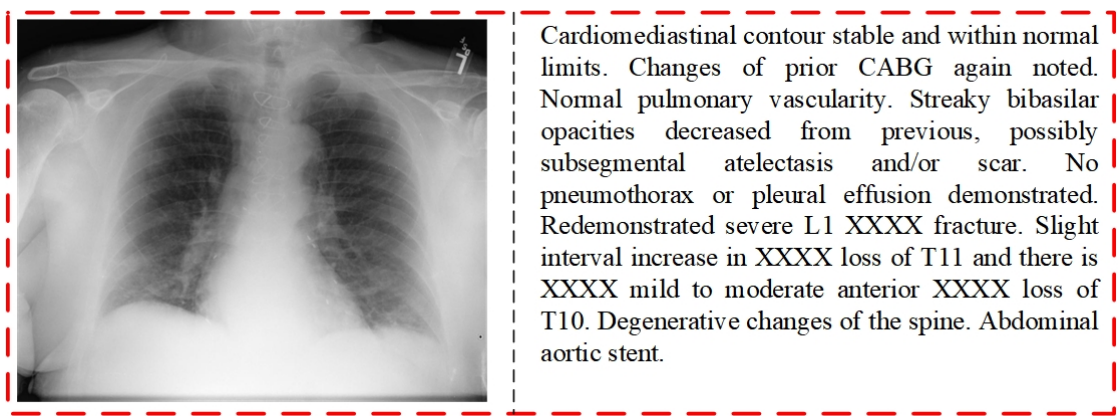

Fig. 1 A representative sample of frequently utilized chest X-ray data.

However, the task of generating diagnostic radiology reports presents significant difficulties in contrast to the conventional image captioning task. (1) Data bias: The size of data from normal reports greatly exceeds that of abnormal reports, leading to a highly imbalanced distribution of positive and abnormal data. Indeed, a significant bias in the data can significantly reduce the effectiveness of data-driven neural networks. To address the negative impact of data bias on performance, Srinivasan et al.[1] developed a multilayer network utilizing a transformer framework for generating reports. Specifically, the architecture comprises a detection network for categorizing normal and abnormal images, a label classification network for generating image labels, and a final report generation network. Similarly, Liu et al.[2] endeavored to address the issue of data bias by emulating the working patterns of radiologists in order to generate accurate diagnostic reports. You et al.[3] introduced the Align-Transformer framework, comprising the Align Hierarchical Attention (AHA) and Multi-Grained Transformer (MGT). The Multi-Head Attention (MHA) module is employed to pinpoint abnormal regions in the image, while the MGT utilizes an adaptive attention mechanism for diagnostic report generation. Li et al.[4] proposed employing contrast learning for the generation of radiology diagnostic reports.

Another significant challenge in medical diagnostic report generation lies in the interpretability of the model. Indeed, the primary reason why generating medical diagnostic reports is largely unacceptable to doctors or patients lies in the inability to explain the model's decisions. Hou et al.[5] proposed the RATCHET transformer network architecture to generate medical diagnostic reports. The architecture includes a visual feature extractor and transformer decoder which utilizes an attention mechanism to achieve semantic alignment between visual and textual features, thus enhancing the model's interpretability. In another work, Chen et al.[6] developed cross-modal memory networks (CMNs) that use supplementary memory matrices to store information linking image and text features. Throughout the encoding and decoding process, memory queries and responses are executed in order to acquire cross-modal shared data.

To address the aforementioned issues, we present a new approach called Dynamic Multi-Domain Knowledge Networks (DMDK). This framework comprises four modules: Chest Feature Extractor (CFE), Dynamic Knowledge Extractor (DKE), Specific Knowledge Extractor (SKE), and Multi-knowledge Integrator (MKI). Precisely, the CFE module is tasked with extracting the initial visual features from the image. In order to address the significant issue of data bias, the DKE dynamically assigns disease topic labels to abnormal image visual features, thereby extracting more effective visual information. Indeed, the DPK module plays a crucial role in extracting $m$ disease topic labels from the diagnostic reports and integrating them with the image features. This process aids in learning the original image features and alleviates the issue of visual bias. The SKE module leverages the disease topic labels extracted by the DKE module to update the knowledge graph. This serves to mitigate the issue of text bias and enhance the model's interpretability. Indeed, knowledge graphs can boost the explanatory capacity of models, aiding physicians and patients in understanding the underlying principles of decision-making. This, in turn, augments the reliability and comprehensibility of both models and systems. The primary responsibility of the MDK module lies in integrating all available knowledge and generating the ultimate diagnostic report.

We conducted comprehensive experiments on two publicly available datasets, IU X-ray and MIMIC-CXR. The experimental results demonstrate the effectiveness of our method. In summary, our main contributions are as follows:

1．We introduce a novel framework that leverages dynamic multi-domain knowledge to produce high-quality diagnostic radiology reports.

2．By constructing dynamic disease topic labels, our approach effectively mitigates the serious visual bias problem in the radiology report generation task.

3．We design a method for dynamically updating the knowledge graph, which effectively reduces text bias and enhances the model's interpretability.

4．We conducted comprehensive experiments on two publicly available datasets. The experimental results affirm the superiority of our approach.

## 2．Related work

### 2．1．Chest X-ray Report Generation

Unlike traditional image captioning tasks, the automatic generation of radiology diagnostic reports demands not only the generation of accurate and coherent diagnostic reports, but also the composition of these reports typically entails lengthy paragraphs to comprehensively describe the clinical symptoms[23][24][25][28][30]. Li et al.[23] introduced a Hybrid Retrieval-Generation Reinforced Agent (HRGR-Agent) approach, which integrates rule-based retrieval with learning-based generation, optimized through reinforcement learning. Specifically, the HRGR-Agent assembles a template database comprising commonly occurring normal utterance templates. During the decoding process, sentence topic states are initially generated, followed by the retrieval strategy module determining whether to retrieve template sentences or autonomously generate new ones. The retrieval strategy module and the generation module undergo joint training using reinforcement learning, utilizing sentence-level and word-level reward functions. Liu et al.[24] introduced a Competence-based Multimodal Curriculum Learning (CMCL) framework, which trains the model by progressively learning from simple to complex samples. They also devised multiple metrics for assessing sample difficulty, considering both visual and textual complexity perspectives. Jing et al.[25] introduced a multi-task learning framework for the simultaneous prediction of labels and generation of diagnostic reports. They also devised a collaborative attention mechanism that effectively concentrates on both the visual information of the image and the predicted semantic labels, enabling accurate localization and description of the abnormal region. Yang et al.[28] developed a triple-branch network (TriNet) to collectively encode visual and semantic features. TriNet integrates visual features, reported semantic features, as well as semantic features derived from disease topic labels, and feeds them into a decoder to generate reports. Similarly， to address the challenge of accurately identifying crucial abnormal regions in the image, Xie et al.[30] introduced an Attention-based Abnormal-Aware Fusion Network (A3FN). This network aims to enhance the model's capability in detecting abnormal regions within the image.

While the aforementioned methods have demonstrated promising results in medical report generation, it's noteworthy that a majority of them rely on the CNN-LSTM architecture. Recently, the CNN-Transformer architecture has emerged as a new paradigm in this field[21][31]. Chen et al. [21] introduced a memory-driven Transformer model, incorporating a relational memory module to capture crucial information during the generation process. Additionally, memory-driven conditional layer normalization is devised to integrate this memory into the Transformer's decoder, thereby enhancing the report generation process. Nguyen et al.[31] proposed an end-to-end multitasking framework for the automatic generation of diagnostic radiology reports. This framework incorporates a classification module designed to extract disease-related features from both medical images and diagnostic reports. A Transformer-based text generation module was constructed, which takes the features extracted by the classification module as input and utilizes them to generate fluent medical reports. Finally, a differentiable interpretation module is employed to assess the coherence of the generated reports with the output of the classification module and refine the generated reports accordingly.

## 2.2. Knowledge Graph

The utilization of knowledge graphs in the domain of diagnostic radiology report generation has steadily gained prominence as a significant research focus. A knowledge graph is a semantic-based representation of knowledge that allows for the visual depiction of entities, attributes, and their relationships in the form of a graph. In the medical domain, Knowledge Graphs can be leveraged to establish a comprehensive medical knowledge base. This resource empowers doctors to swiftly access pertinent information about diseases, symptoms, treatment options, and more, ultimately leading to heightened efficiency and precision in their diagnostic processes.

Zhang et al.[12] introduced a method that integrates a knowledge graph with medical image report generation. More precisely, they formulated a knowledge graph encompassing 20 prevalent thoracic lesion categories, which was utilized as prior knowledge input to a deep neural network. Liu et al.[2] introduced a Posterior-and-Prior Knowledge Exploring-and-Distilling approach (PPKED) for medical diagnostic report generation. PPKED acquires prior medical knowledge by creating a knowledge graph containing the most prevalent types of abnormalities. All the aforementioned studies rely on a static set of disease topic labels for knowledge graph construction. In our research, we propose an approach capable of dynamically updating the knowledge graph to create a tailored knowledge graph for each medical image. This dynamic knowledge graph is more adaptable to real-world scenarios and can enhance the explanatory capacity of the generated reports.

## 3. Our Proposed

In this section, we provide a comprehensive introduction to our proposed DMDK network. DMDK comprises the following key components: Chest Feature Extractor (CFE), Dynamic Knowledge Extractor (DKE), Specific Knowledge Extractor (SKE), and Multi-knowledge Integrator (MKI) module, as illustrated in Fig. 2. We initiate with an overview of the DMDK network, followed by an introduction to the foundational model utilized in this paper. Subsequently, we offer an in-depth exposition of each module within the MDDK framework.

## 3.1. Overview

The process of radiologists composing radiology diagnostic reports typically entails a thorough examination of abnormal areas in the images, followed by the integration of their medical expertise and professional experience to finalize the reports. Drawing inspiration from this operational paradigm, we incorporated posteriori knowledge (disease topic labels) and a priori knowledge (knowledge graph) to support the DMDK network in the generation of diagnostic reports. The structure of the DMDK network is illustrated in Fig. 2. During the training phase, the DMDK processes one medical image at a time, along with the diagnostic report and the knowledge graph as inputs, subsequently generating the corresponding radiology report. In particular, the left segment of Fig. 2 illustrates the Chest Feature Extractor (CFE) module, which is based on the pre-trained ResNet152. The original visual features extracted by the CFE are fused as query matrices into both the DKE and SKE modules, individually, with the aim of mitigating the pronounced data bias issue. We will provide a detailed description of the CFE module in Section 3.2.

Next are two knowledge fusion modules DKE and SKE. The DKE module is responsible for extracting disease topic labels from radiology reports and fusing them with image features to enhance the original visual features. Disease topic labels encompass not only common disease descriptors but also convey essential information about the image. Consequently, the model is equipped not only to associate anomalous regions with pertinent disease topic label information but also to acquire supplementary semantic insights. This mirrors the workflow of radiologists, who assign pertinent disease topic labels to abnormal regions during image examination. This approach significantly enhances the model's capacity to identify abnormal regions within an image. Furthermore, within the PSK module, we dynamically construct a specific knowledge graph for each sample. Due to its wealth of semantic and structural information, the knowledge graph serves a dual purpose. It not only aids in alleviating substantial textual data bias but also elevates the interpretability of the report generation process. We will provide a comprehensive description of the DPK module in Section 3.3 and the SPK module in Section 3.4.

Following the paradigm of diagnostic radiology report generation, the MDK module employs a standard transformer decoder for report generation. To enhance the quality of the generated reports, we incorporated the knowledge extracted from the individual modules into the decoder to obtain the fused visual features. The decoder employs the fused visual features for radiology report generation. More detailed information will be presented in Section 3.5.

***Basic Module*** We adopt the base model proposed by Vaswani et al.[13], which incorporates Multi-Head Attention (MHA) and Feed-Forward Network (FFN). The Multi-Head Attention (MHA) is composed of $n$ parallel heads, with each head being defined as scaled dot-product attention:

$$\text{Att}_i(X, Y) = softmax(\frac{XW_i^Q(YW_i^K)^T}{\sqrt{d_k}})YW_i^V$$

$$\text{MHA}(X, Y) = concat(\text{Att}_1(X, Y), \text{Att}_2(X, Y), ..., \text{Att}_h(X, Y))W^O$$

where $X \in \mathbb{R}^{l_x \times d}$ and $Y \in \mathbb{R}^{l_y \times d}$ represent the query matrix and the Key/Value matrix, respectively. $W_i^Q \in {}^{d \times d_n}$, $W_i^k \in {}^{d \times d_n}$, $W_i^V \in {}^{d \times d_n}$ and $W^O \in {}^{d \times d}$ are the parameter matrices, where $d_n = d/n$ and $concat$ stands for concatenation operation.

Similarly, The FFN is defined as follows:

$$\text{FFN}(x) = \max(0, x\text{W}_f + \text{b}_f)\text{W}_{ff} + \text{b}_{ff}$$

where $\max(0, \cdot)$ represents the ReLU activation function. $\text{W}_f \in \mathbb{R}^{d \times 4d}$ and $\text{W}_{ff} \in \mathbb{R}^{4d \times d}$ denote learnable matrices for linear transformation. $\text{b}_f$ and $\text{b}_{ff}$ are bias terms.

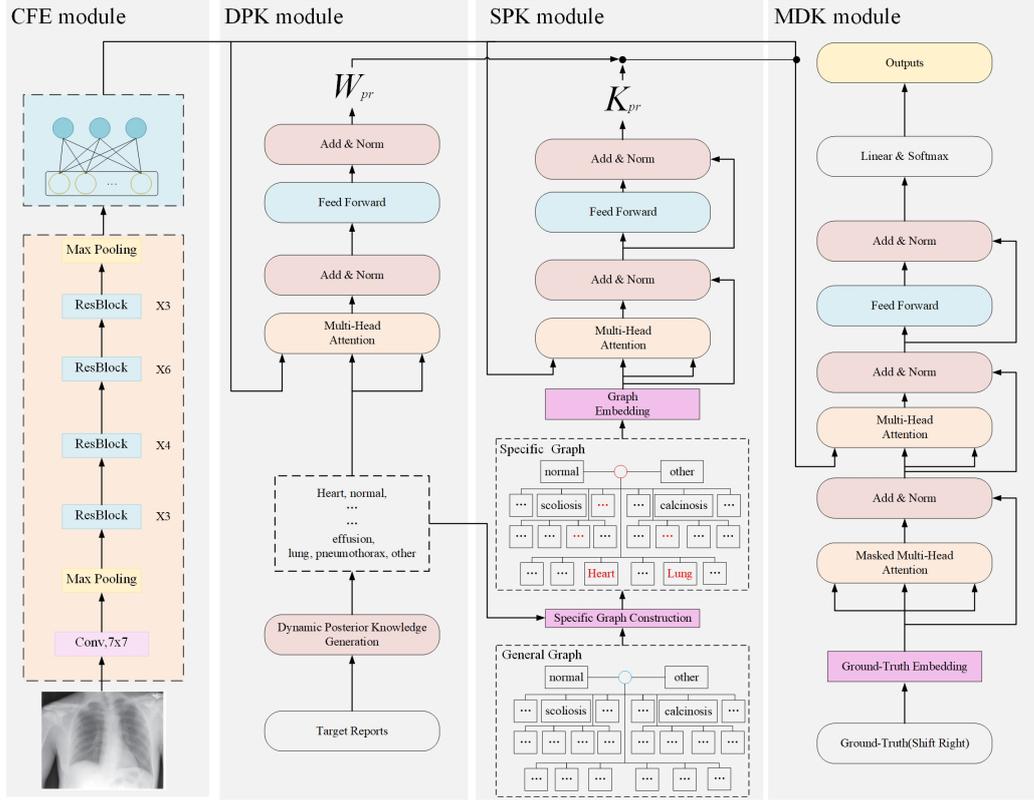

Fig. 2 Network framework.

## 3.2. Chest Feature Extractor (CFE)

In essence, the diagnostic report generation task is to implement domain adaptation learning from the image domain to the text domain. Given any medical image $I$, its original visual features can be defined as $X=\{x_1, x_2, ..., x_s, ..., x_N\}$, where $x_s \in \mathbb{R}^{N \times d}$ is the visual feature vector extracted from the radiology image by the feature extractor, and $d$ represents the dimension of the feature vector. Similarly, the actual diagnostic report corresponding to the image can be defined as $Y=\{y_1, y_2, ..., y_m, ..., y_M\}$, where $y_m \in T$ are the generated tokens, $M$ represents the length of the diagnostic report, $T$ is the vocabulary of all possible tokens.

Following[6][12][21], we adopt the convolutional neural network Resnet152[26] as the chest feature extractor. Specifically, the Resnet152 extracts $2,048$ $7 \times 7$ image features which are further projected into $1536$ $7 \times 7$ image features, resulting $X=\{x_1, x_2, ..., x_s, ..., x_N\} \in \mathbb{R}^{N_1 \times d}$ ($N_1 = 49, d = 1536$). The above process can be formalized as:

$$X=f(CNN(I))$$

where $CNN$ stands for the Resnet152 and the $f(\cdot)$ is the mapping function.

## 3.3. Dynamic Knowledge Extractor(DKE)

While prior knowledge has demonstrated its effectiveness in various diagnostic report generation tasks[2][22], it is often assumed to be a static or fixed set of information. Given that static information collections struggle to adapt to changing data, we have introduced the biomedical named entity recognition model, Stanza, proposed by Zhang et al.[36], to analyze the entity type of each disease topic label in radiology diagnostic reports. Subsequently, we construct dynamic disease topic tags based on the identified entity types.

The Stanza model contains a total of five types of entities: ***anatomy, observation, anatomy modifier, observation modifier and uncertainty***. To mitigate the potential increase in noise resulting from the introduction of disease topic labels, we refrain from utilizing all entities predicted by the Stanza model as disease topic labels. In a specific process, we initially locate the first entity with the type "***anatomy***" from the entity sequence. Subsequently, we retrieve an entity with a type other than "***anatomy***" from the entity sequence. The algorithmic flow is illustrated in Tab. 1. Exactly, we contend that each entity categorized as "***anatomy***" should be matched with a "***non-anatomy***" entity to create a set of labels with distinct and well-defined meanings. For instance, consider the entity "***heart***" with the type "***anatomy***" and the entity "***cardiomegaly***" with the type "***observation***". These entities can be combined to form a pair of tags < "***heart***", "***cardiomegaly***">, which carries a clear and meaningful interpretation. Moreover, to ensure that disease topic labels are not empty, we employ the graph entities outlined in[12] as the foundational disease topic labels for our approach.

Tab. 1 Disease Topic Labels Search

| **Algorithm 1:** Disease Topic Labels Search |
|---|
| **Input:** The radiology report *R* pre-retrieved from the training corpus.<br>**Output:** Entities and their types for each word in the radiology report *R*.<br>  1.    *entity* ← *stanza*(*R*)       // The type of *entity* is array.<br>  2.    *type* ← *stanza*(*R*)        // The type of *type* is array.<br>  3.    *flag* ← False<br>  4.    *temp* ← "ANATOMY" |

```
5.    i ← 0
6.    j ← length(entity)
7.    k ← 0
8.    while i ≤ j  do
9.        if type [i] == temp and j-i != 1   then
10.           if type [i+1] != temp   then
11.               tags[k++] ← entity[i]
12.               flag ← True
13.       if flag == True and entity_type [i] != "ANATOMY"   then
14.           tags[k++] ← entity[i]
15.           flag ← False
16.   return  tags
```

After obtaining the disease topic labels, we encode them and achieve multi-modal feature alignment with the visual features of the image through an attention mechanism. Indeed, we employ the image feature $X$ as the query matrix and the disease topic label $tags$ as the Key and Value matrix for attention calculation, resulting in the enhanced visual information denoted as $W'$:

$$W = Embedding(tags)$$

$$W' = MHA(X, W)$$

where, $tags$ represents disease topic label, $W' \in \mathrm{R}^{N_1 \times d}$ 。

## 3.4. Specific Knowledge Extractor(SKE)

Knowledge graphs have been extensively validated and applied in tasks related to medical report generation. Zhang et al.[12] proposed utilizing a static Chest Knowledge Graph $G_{st}$ as prior knowledge to enhance crucial disease-related feature information and generate diagnostic reports. $G_{st}$ comprises 27 entities, along with a root node $V$ representing global features, and an adjacency matrix $A = \{E_{ij}\}$ representing edges. Each node represents a disease topic label, and $E_{ij}$ is set to 1 when there is an association between two nodes. Indeed, because $G_{st}$ is a knowledge graph constructed using a predetermined set of disease topic labels, it will not be effectively updated during training. Hence, $G_{st}$ exhibits several noteworthy limitations: (1) Given the pronounced data bias issues in the domain of diagnostic report generation, $G_{st}$ does not encompass all disease topic labels in the dataset. As shown in the red part in Fig. 3, the disease topic label "trachea" is absent from the general knowledge graph. (2) Indeed, due to the diversity of provided medical images, it is challenging to comprehensively represent feature information using a standardized static knowledge graph. To address the aforementioned issue, we propose utilizing dynamic disease topic labels to construct a specific Knowledge Graph, as illustrated in Fig. 4. After obtaining the specific Knowledge Graph $G_{sp}$, we fuse it with the original visual features to elevate the quality of the generated reports.

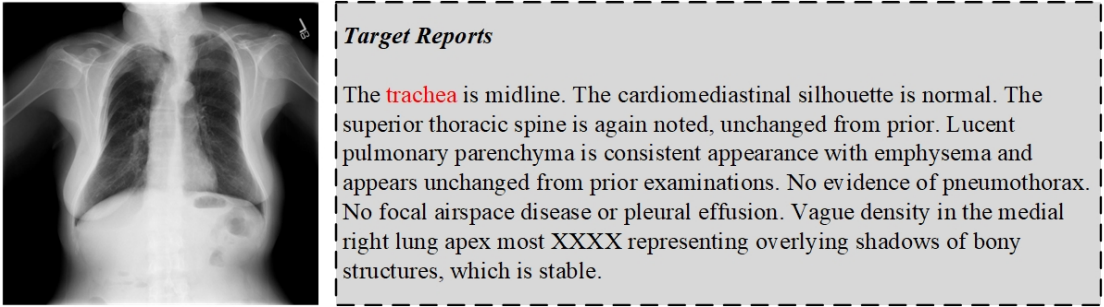

Fig. 3 An illustration of one sample from IU X-ray.

**Specific Graph Construction** We employ a top-down approach to construct the knowledge graph. In particular, the foundational structure is initially established using general knowledge, and subsequently, disease topic tags are employed to introduce nodes or redefine the relationships between nodes. We utilize $G_{st}$ as our basic knowledge graph, which contains 28 entities that are composed of a root node representing all entities, 7 organs and tissues, and 20 disease keywords, as shown in the general graph part of Fig. 4. $G_{st}$ represents a finding and is denoted by a disease topic label. Except for "normal", "other" and "Foreign Object" all disease topic labels are linked according to the associated body organs.

Tab. 2 Specific Graph Construction

| **Algorithm 2:** Specific Graph Construction |
|---|
| **Input:** The radiology report *R* pre-retrieved from the training corpus. <br> **Output:** Entities and their types for each word in the radiology report *R*. <br>   1.    *entity* ← *stanza*(*R*)        // The type of *entity* is array. <br>   2.    *type* ← *stanza*(*R*)         // The type of *type* is array. <br>   3.    *flag* ← False <br>   4.    *temp* ← "ANATOMY" |

|     |     |
| --- | --- |
| 5.  | $i \leftarrow 0$ |
| 6.  | $j \leftarrow$ length(*entity*) |
| 7.  | $k \leftarrow 0$ |
| 8.  | **while** $i \leq j$ **do** |
| 9.  |    **if** *type* [i] == *temp* and *j*-i != 1   **then** |
| 10. |       **if** *type* [i+1] != *temp*   **then** |
| 11. |          *source*[k++] $\leftarrow$ *entity*[i] |
| 12. |          *flag* $\leftarrow$ True |
| 13. |    **if** flag == True and *entity_type* [i] != "ANATOMY"   **then** |
| 14. |       *target*[k++] $\leftarrow$ *entity[i]* |
| 15. |       *relations*[k] $\leftarrow$ *type[i]* |
| 16. |       *flag* $\leftarrow$ False |
| 17. | **return**   *source, target, relations* |

To obtain a specific knowledge graph $G_{sp}$ for each given medical image $X$, we use the disease topic labels *tags* (which contain m entities) predicted by the DKE module to update the generic knowledge graph $G_{st}$. Similar to constructing disease topic labels, the nodes and edges of the knowledge graph are predicted by the Stanza model. The final nodes and edges of the graph are determined according to the algorithm outlined in Tab. 2. Next, we select one entity $e_i$ from the disease topic labels *tags* at a time to update $G_{st}$. If $e_i$ is not in $G_{st}$ and is associated with other entities $e_j$, it is added to $G_{sp}$. We iterate over all the entities in disease topic labels *tags* and repeat the above process to obtain the specific knowledge graph $G_{sp}$. We employ the five entity relationships proposed by Stanza to describe the relationship between source and target entities. Through the aforementioned top-down approach, a dynamic knowledge graph can be constructed for each provided medical image. Finally, the graph convolutional neural network[34] serves as a knowledge graph encoder to extract features $M$.

Similar to the approach detailed in Section 3.3, we employ an attention mechanism to align the knowledge graph and image visual features, creating multi-modal features that enhance the representational capacity of the original image features:

$$M = GCN(G_{sp})$$

$$M' = MHA(X, M)$$

where $M$ represents the encoded features of the knowledge graph. $GCN$ is a graph

convolutional neural network (GCN) [34]。

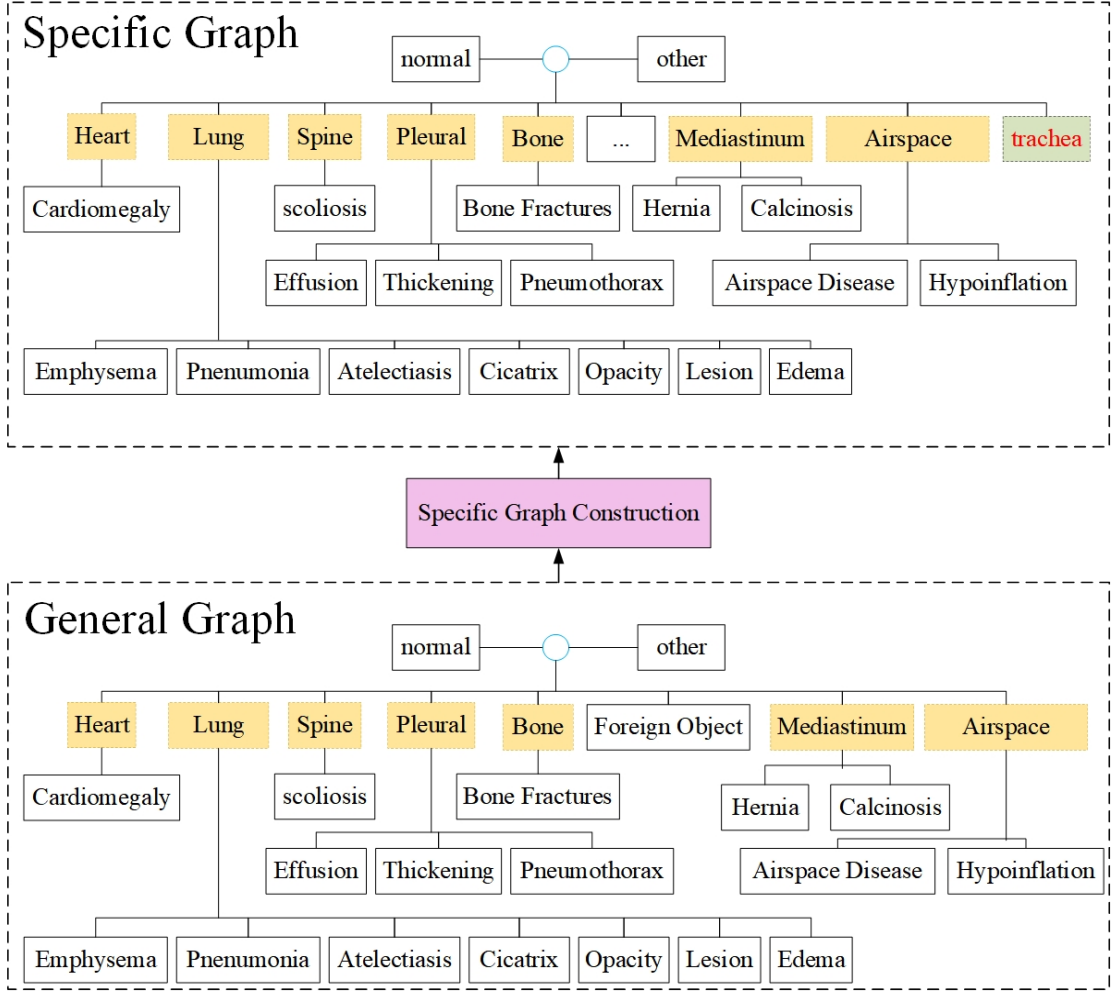

Fig. 4 The specific knowledge graph

## 3.5. Multi-knowledge Integrator(MKI)

Following the paradigm of radiology report generation, we employ a standard Transformer decoder to generate the final report. After receiving the enhanced visual features $W'$ and $M'$, at each time step $h$, the MKI module takes the encoding of the current input word $y_h = t_h + e_h$ as input. The entire decoding process can be formalized as follows:

$$x_h = \text{MHA}(y_h, y_{1:h})$$

The objective of our approach is to generate diagnostic reports utilizing medical imaging features $I$, and to augment the performance of network-generated reports by incorporating disease topic labels $W'$ and dynamic knowledge graphs $M'$ information. Therefore, we

employ MAA to embed $W'$ and $M'$ into the network to assist in the generation process:

$$X' = MHA(X, \lambda_1 X + \lambda_2 W' + \lambda_3 M')$$

$$x'_h = MAA(y_h, X', W', M')$$

where $\lambda_1, \lambda_2, \lambda_3 \in (0,1)$ and $\lambda_1 + \lambda_2 + \lambda_3 = 1$.

Finally, $x'_h$ is passed through a Feed-Forward Network (FFN) and a Multi-Layer Perceptron (MLP) to predict the next word:

$$p_h = \text{softmax}(\text{FFN}((x'_h)W_p + b_p)$$

where $W_p$ and $b_p$ represent learnable parameters and bias terms, respectively.

## 4. Experiments

In this section, we first introduce the two datasets used in the experiments. Next, we present some widely used evaluation metrics and baselines. Finally, we provide an analysis of our proposed method.

### 4.1. Datasets, Metrics, Baselines and Settings

#### 4.1.1 Datasets

We conducted experiments on two extensively employed datasets, IU X-ray[15] and MIMIC-CXR[16], to validate the effectiveness of our proposed model. The IU X-ray dataset is a publicly accessible dataset designed for medical image analysis, specifically focusing on X-ray images and facilitating the evaluation of radiology diagnostic report generation. The dataset, provided by Indiana University, comprises a total of 7,470 chest X-ray images and 3,955 corresponding radiology diagnostic reports. The MIMIC-CXR dataset is currently the largest radiology imaging dataset available. It comprises a total of 473,057 chest X-ray images and 206,563 associated radiology diagnostic reports.

For both datasets, we pre-processed radiology reports by tokenizing, converting to lowercase, and filtering words with a frequency of less than 3. The dataset is partitioned into training, validation, and test sets using a ratio of 7:1:2. The training set, validation set, and test set are mutually exclusive with no overlap. Finally, we obtained 2,069/270,790 training samples, 296/2,130 validation samples, and 590/3,858 test samples on the IU X-ray and MIMIC-CXR datasets respectively. The data statistics are presented in Tab. 3.

Tab. 3 Number of IU X-ray and MIMIC-CXR data sets used in this study. Notably, each report within the IU X-ray dataset comprises two medical images.

| Datasets | Type | Total | Train | Val | Test |
| --- | --- | --- | --- | --- | --- |
| IU X-ray | Images | 5,910 | 4,138 | 592 | 1,180 |
| | Reports | 2,955 | 2,069 | 296 | 590 |
| MIMIC-CXR | Images | 276,778 | 270,790 | 2,130 | 3,858 |
| | Reports | 276,778 | 270,790 | 2,130 | 3,858 |

### 4.1.2 Metrics

We define the generated sentences as reference translations, while the real sentences serve as candidate translations. Following the standard paradigm of image captioning, we employ the widely used evaluation metrics including BLEU[17], ROUGE-L[19], and CIDEr[20] for assessing the performance of our model.

BLEU (Bilingual Evaluation Understudy) is a metric used to evaluate the quality of translations. BLEU is calculated based on N-gram overlap, which assesses the quality of translation by measuring the shared N-grams (N consecutive words) between the reference translation and the candidate translation. The BLEU score is calculated as follows:

$$P_n = \frac{\text{Total matching N-grams}}{\text{Total N-grams in candidate translation}}$$

Here, the total number of N-grams refers to the N-grams present in the candidate translation.

$$BP = \begin{cases} 1 & \text{if } c > r \\ e^{(1-r/c)} & \text{if } c \leq r \end{cases}$$

$$BLEU = BP * \exp(\sum_{n=1}^{N} W_n \log P_n)$$

where $BP$ stands for brevity penalty. It is a factor used to penalize translations that are shorter than the reference translations. $r$ is the length of the closest reference translation. $c$ is the length of the candidate translation. $W_n$ refers to the weight assigned to the precision of N-grams.

ROUGE-L (Recall-Oriented Understudy for Gisting Evaluation) is one of the metrics in the ROUGE family used for automatic evaluation of tasks like text summarization and machine translation. Specifically, ROUGE-L is based on the Longest Common Subsequence (LCS) to measure the similarity between a reference summary and a generated summary.:

$$ROUGE-L = \frac{(1+\beta^2)R_{lcs}P_{lsc}}{R_{lcs}+\beta^2 P_{lsc}}$$

$$R_{lcs} = \frac{LSC(X,Y)}{m}$$

$$P_{lsc} = \frac{LSC(X,Y)}{n}$$

where $X$ represents the candidate translation, $Y$ represents the reference translation, $LSC(X,Y)$ is the length of the longest common subsequence of the candidate translation and the reference translation, and $m$ and $n$ represent the length of the reference translation and the length of the candidate translation, respectively.

CIDEr (Consensus-based Image Description Evaluation) is an evaluation metric commonly used in the field of computer vision and natural language processing, especially for tasks related to image captioning. CIDEr evaluates the quality of generated image captions by considering both consensus and diversity in the generated captions. It measures how well a generated caption aligns with multiple reference captions for a given image.:

$$g_k(s_{ij}) = \frac{h_k(s_{ij})}{\sum_{wl\in\Omega} h_l(s_{ij})} \log(\frac{|I|}{\sum_{I_p\in I} \min\{1,\sum_q h_k(s_{pq})\}})$$

$$CIDEr_n = \frac{1}{m}\sum_j \frac{g^n(c_i)^T g^n(s_{ij})}{\|g^n(c_i)\|\bullet\|g^n(s_{ij})\|}$$

$$CIDEr(c_i, S_i) = \sum_{n=1}^{N} w_n CIDEr_n(c_i, S_i)$$

where $\Omega$ represents all the words of all n-grams, $I$ represents the number of all images in the dataset, $(\cdot)$ represents the number of times $w$ appears in the candidate sentence $c_i$ or the reference sentence $s_{ij}$, and $w_n$ represents different N-grams.

### 4.1.3 Baselines

We compared the proposed method with previous state-of-the-art models, namely R2Gen[21]，PPKED[2]，CA[9]，AlignTransformer[3]，MKG[12]，MGSK[22]，HRGR[23]，CMCL[24]，CoAtt[25]，CMN[6], ASGMD[35]. The details are as follows:

(1) R2Gen proposed a memory-driven transformer model to generate radiology reports. Specifically, R2Gen designed a relational memory module to record critical information in the previous generation process. It proposed a new memory-driven conditional layer Normalization method to integrate the relational memory module into the transformer decoder.

(2) PPKED introduced the Posterior Knowledge Explorer (PoKE) and Priori Knowledge Explorer (PrKE) to address the issue of bias in both visual and textual data.。

(3) CA suggested a contrastive attention model which aims to capture the visual features of abnormal regions. It achieves this by comparing input images with established normal images, aiding the model in providing more accurate descriptions of chest X-ray abnormalities.

(4) AlignTransformer proposed the Align Hierarchical Attention module to predict disease labels. Subsequently, it employed hierarchical alignment between visual regions and disease labels to acquire multi-granularity feature representations, enabling a concentrated focus on abnormal areas.

(5) MKG proposed the integration of a knowledge graph into the automated generation of chest X-ray reports, aiming to enhance the accuracy and relevance of the generated reports.

(6) MGSK proposed a novel framework for radiology report generation that integrates both general knowledge and specific knowledge.

(7) HRGR-Agent introduced a method called Hybrid Retrieval-Generation Reinforced Agent for the automatic generation of medical image reports.

(8) CMCL introduced multiple sample difficulty evaluation metrics that take into account visual complexity and text complexity to assess the challenge of accurately describing anomalies.

(9) CoAtt introduced a multi-task learning framework along with a collaborative attention mechanism. This framework enables simultaneous prediction of tags and generation of medical reports, addressing the challenge of generating heterogeneous information.

(10) CMN introduced a cross-modal memory network utilizing a shared memory matrix to store the correspondence between images and text, thus enhancing the interaction between the two modalities.

(11) ASDMN proposed an auxiliary signal-guided and memory-driven network for the automatic generation of medical imaging reports.

### 4.1.4 Settings

All experiments in this paper are conducted under the following configuration: Intel(R) Xeon(R) Platinum 8358 CPU @ 2.60GHz, 2.00 TB RAM, and eight Nvidia GeForce GTX 1080 Ti GPU。We utilize the pre-trained Resnet[26] as the chest feature extractor. The feature maps extracted by the Resnet 152 are of size $2,048 \times 7 \times 7$, which are further mapped to $1536 \times 49$. It is noteworthy that, in the case of the IU X-ray dataset, the chest feature extractor takes two CT images of the patient as input simultaneously. Based on the model's performance on the validation set, we established different hyperparameters for training on the IU X-ray and MIMIC-CXR datasets, respectively. Specifically, the IU X-ray dataset and MIMIC-CXR dataset utilized initial learning rates of 1e-4 and 5e-5, respectively. The batch size for both datasets is set to 128.To prevent overfitting, the L2 regularization coefficient is set to 0.001. The model employs 8 heads and has a dimension of 512 for the multi-head attention. The ADAM optimizer is utilized to minimize the cross-entropy loss function.

### 4.2. Quantitative results and Analysis

Tab. 4 and Tab. 5 show the performance of our proposed model and previous advanced models on the IU X-ray and MIMIC-CXR datasets, respectively. In line with the natural language generation paradigm, our primary evaluation metrics consist of BLEU, CIDEr, and ROUGE-L.

First and foremost, our model outperforms all other radiology report generation models in performance. Compared to the state-of-the-art model MGSK[22] on the IU X-ray dataset, our model has improved BLEU-1, BLEU-2, BLEU-3, BLEU-4, and CIDEr from 0.496, 0.327, 0.238, 0.178, 0.382 to 0.520, 0.357, 0.264, 0.201, 0.430, respectively. The average improvement is 10.08%. Additionally, compared with the state-of-the-art ASGMD model[35], the ROUGE-L performance shows an improvement from 0.397 to 0.414. Compared to the state-of-the-art model AlignTransformer[3] on the MIMIC-CXR dataset, our model has increased the BLEU-1, BLEU-2, BLEU-3 evaluation metrics from 0.378, 0.235, 0.156 to 0.617, 0.504, 0.394. The average improvement amounts to 0.249. Compared to the MGSK model[22], BLEU-4 and CIDEr increased from 0.115 and 203 to... respectively. In addition, compared to the ASGMD model[35], ROUGE-L improved from 0.286 to...The experimental results demonstrate the effectiveness of our proposed model, showcasing that leveraging dynamic prior knowledge significantly enhances the performance of radiology diagnostic report generation. An interesting discovery is that the effectiveness of directly employing a model designed for natural image captioning to generate radiology diagnostic reports (as demonstrated in rows 1/2/3 in Tab. 4 and Tab. 5) is significantly lower compared to using a specialized radiology report generation model (as illustrated in Tab. 4 and elsewhere in Tab. 5). This highlights the crucial significance of constructing generative models tailored specifically for radiology reporting. It's important to note that the current assessment of model performance typically leans towards BLEU-4, meaning the best BLEU-4 result is chosen for comparison. However, since different metrics evaluate the impact of different aspects of the model, we also provide the best performance that our model can achieve on each metric. Indeed, our model demonstrates superior performance on the CIDEr and ROUGE-L evaluation metrics, as shown in Tab. 6. Given the diverse considerations of downstream tasks, it's essential to choose an appropriate model for pre-training.

Tab. 4 Performance Comparison between Our Model and Baseline Models on the IU X-ray Dataset. The optimal and suboptimal results are highlighted in bold and underlined, respectively. * indicates the result after we reproduced it using the code they released. The rest of the results are quoted from the original article. BLEU-n evaluates the accuracy of generated radiology reports. CIDEr verifies whether the generated radiology report aligns with the crucial information in the real report. ROUGE-L measures the recall of generated radiology reports. A larger value for all evaluation metrics indicates better performance of the model. The performance results of our model are derived from the average of five experimental results.

| Model | BLEU-1 | BLEU-2 | BLEU-3 | BLEU-4 | CIDEr | ROUGE-L |
|---|---|---|---|---|---|---|
| S&T* | 0.216 | 0.124 | 0.087 | 0.006 | 0.294 | 0.306 |
| SA&T* | 0.399 | 0.251 | 0.168 | 0.118 | 0.302 | 0.323 |
| AdaAtt* | 0.220 | 0.127 | 0.089 | 0.068 | 0.295 | 0.308 |
| R2Gen | 0.470 | 0.304 | 0.219 | 0.165 | - | 0.371 |
| PPKED | 0.483 | 0.315 | 0.224 | 0.168 | - | 0.351 |
| CA | 0.492 | 0.314 | 0.222 | 0.169 | - | 0.381 |
| AlignTransformer | 0.484 | 0.313 | 0.225 | 0.173 | - | 0.379 |
| MKG | 0.441 | 0.291 | 0.203 | 0.147 | 0.304 | 0.367 |
| MGSK | <u>0.496</u> | <u>0.327</u> | <u>0.238</u> | <u>0.178</u> | <u>0.382</u> | 0.381 |
| HRGR | 0.438 | 0.298 | 0.208 | 0.151 | 0.343 | 0.322 |
| CMCL | 0.473 | 0.305 | 0.217 | 0.162 | - | 0.378 |
| CoAtt | 0.455 | 0.288 | 0.205 | 0.154 | - | 0.277 |
| CMN | 0.475 | 0.309 | 0.222 | 0.170 | - | 0.375 |
| ASGMD | 0.489 | 0.326 | 0.232 | 0.173 | | <u>0.397</u> |
| Ours | **0.520** | **0.357** | **0.264** | **0.201** | **0.430** | **0.414** |

Tab. 5 Performance Comparison between Our Model and Baseline Models on the IU X-ray Dataset. The optimal and suboptimal results are highlighted in bold and underlined, respectively. * indicates the result after we reproduced it using the code they released. The rest of the results are quoted from the original article. BLEU-n evaluates the accuracy of generated radiology reports. CIDEr verifies whether the generated radiology report aligns with the crucial information in the real report. ROUGE-L measures the recall of generated radiology reports. A larger value for all evaluation metrics indicates better performance of the model. The performance results of our model are derived from the average of five experimental results.

| Model | BLEU-1 | BLEU-2 | BLEU-3 | BLEU-4 | CIDEr | ROUGE-L |
|---|---|---|---|---|---|---|
| S&T* | 0.256 | 0.157 | 0.102 | 0.070 | 0.063 | 0.249 |

| Model | BLEU-1 | BLEU-2 | BLEU-3 | BLEU-4 | CIDEr | ROUGE-L |
|---|---|---|---|---|---|---|
| SA&T* | 0.304 | 0.177 | 0.112 | 0.077 | 0.083 | 0.249 |
| AdaAtt* | 0.311 | 0.178 | 0.111 | 0.075 | 0.084 | 0.246 |
| R2Gen | 0.353 | 0.218 | 0.145 | 0.103 | - | 0.277 |
| PPKED | 0.360 | 0.224 | 0.149 | 0.106 | - | 0.284 |
| CA | 0.350 | 0.219 | 0.152 | 0.109 | - | 0.283 |
| AlignTransformer | 0.378 | 0.235 | 0.156 | 0.112 | - | 0.283 |
| MGSK | 0.363 | 0.228 | 0.156 | 0.115 | 0.203 | 0.284 |
| CMCL | 0.334 | 0.217 | 0.140 | 0.097 | - | 0.281 |
| CMN | 0.353 | 0.218 | 0.148 | 0.106 | - | 0.278 |
| ASGMD | 0.372 | 0.233 | 0.154 | 0.112 | - | 0.286 |
| Ours | | | | | | |

Tab. 6 "ours" represents the best performance our model can achieve, guided by the BLEU-4 metric. "Best" represents the highest performance achievable by our model on each metric.

| Dataset | Dimension | BLEU-1 | BLEU-2 | BLEU-3 | BLEU-4 | CIDEr | ROUGE-L |
|---|---|---|---|---|---|---|---|
| IU X-ray | Best | **0.522** | **0.377** | 0.264 | 0.201 | **0.447** | **0.421** |
| | Ours | **0.520** | **0.357** | **0.264** | **0.201** | **0.430** | **0.414** |
| MIMIC-CXR | Best | | | | | | |
| | Ours | | | | | | |

## 4.3. Qualitative results and Analysis

In this section, we will qualitatively analyze the visualization results of the model. Fig. 5 illustrates the performance comparison between radiology diagnostic reports generated by our method and the state-of-the-art model R2Gen[21] on the IU X-ray dataset. Fig. 5 presents four components of information. The initial section features the target X-ray image. The second section displays the corresponding actual diagnostic report for the X-ray. The third and fourth sections showcase the diagnostic reports generated by the advanced model R2Gen and our model, respectively. The experimental results demonstrate that, in comparison to R2Gen, which is only capable of generating isolated words, our model produces radiology diagnostic reports with a significantly higher degree of overlap with ground truth (as depicted in columns 2-4 of Fig. 5). Furthermore, our model is able to generate reports that completely overlap with real diagnostic reports (shown in column 5 of Fig. 5).

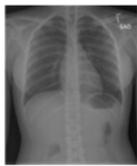

Fig. 5 Performance comparison of radiology diagnostic reports generated by our model and the state-of-the-art model R2Gen[21]. Blue and red fonts indicate the fit of the R2Gen model and our model to ground truth radiology reports, respectively.

## 4. 4. Ablation Study

To evaluate the effectiveness of each proposed module, we conducted ablation experiments on the model using the same set of parameters. We use the standard transformer model as our basic model and verify the effectiveness of each module through superposition. Additionally, to validate the efficacy of our proposed dynamic knowledge, we substituted a set of static disease topic tags and a fixed knowledge graph in its place. The results are presented in Tab. 7 and Tab. 8.

**Effectiveness of DKE** Tab. 7 demonstrates the effectiveness of the DKE module. Compared to the basic model, the performance of BLEU-1, BLEU-2, BLEU-3, and BLEU-4 increased from 0.372, 0.239, 0.172, and 0.129 to 0.440, 0.286, 0.206, and 0.156 respectively. The average improvement is 49.6%. Additionally, the CIDEr and ROUGE-L evaluation metrics increased from 0.365 and 0.348 to 0.378 and 0.378 respectively. The average growth rate is 44.58%. In prior research[35], static knowledge was exclusively embedded into the model to aid in the generation of radiology reports. Hence, to assess the effectiveness of our proposed dynamic disease topic labels, we substitute the disease topic labels extracted by the DKE module with the fixed auxiliary signal proposed by [35]. Tab. 8 illustrates the effectiveness of dynamic knowledge. Compared with fixed topic labels, our model improves the performance of BLEU-1, BLEU-2, BLEU-3, and BLEU-4 from 0.388, 0.264, 0.153, and 0.108 to 0.440, 0.286, 0.206, and 0.156, respectively. In addition, the CIDEr and ROUGE-L evaluation indicators increased from 0.334 and 0.357 to 0.378 and 0.378, respectively. The average growth rate is 9.53%.

**Effectiveness of SKE** Tab. 7 also shows the effectiveness of the SKE module. Specifically, compared to the Base model, with the introduction of the dynamic knowledge graph, the performance metrics of BLEU-1, BLEU-2, BLEU-3, and BLEU-4 increased from... to... . On average, there is an improvement of.....The CIDEr and ROUGE-L evaluation metrics showed improvements from... to... respectively. On average, there is a growth of...In addition, to evaluate the impact of dynamic and static knowledge graphs on model performance, we replaced the SKE module proposed in this study with the knowledge graph constructed by Liu et al.[2]. The results are shown in Tab. 8. Experimental results demonstrate the effectiveness of the dynamic knowledge graph proposed in this study in the task of generating radiology diagnostic reports. Specifically, the performance of BLEU-1, BLEU-2, BLEU-3, and BLEU-4 has improved from [previous scores] to [new scores] respectively. In addition, the CIDEr and ROUGE-L evaluation indicators improved from [previous scores] to [new scores], with an average growth of [average growth percentage]。

Furthermore, to provide a more intuitive explanation of the effectiveness of each module, we conducted an analysis of the radiology diagnostic reports generated by each module. Fig. 6 presents a comparison between the radiology diagnostic reports generated by each individual module and the overall performance of our proposed model. Specifically, compared with the radiology diagnosis report generated by the Base model, after the introduction of dynamic disease topic tags and knowledge graphs (columns 4 and 5 of Fig. 6), the overlap between the report generated by the model and the ground truth report is higher. Compared to the DKE model

(column 4 of Fig. 6), the SKE model (column 5 of Fig. 6) demonstrates superior performance in generating radiology diagnostic reports. This illustrates that constructing a dynamic knowledge graph can enhance the performance of radiology diagnostic reporting tasks.

Tab. 7 We performed ablation experiments on our proposed method using the IU X-ray and MIMIC-CXR datasets. The base model represents our method implemented on the standard transformer model. "+" represents the combination of proposed modules. Bold represents the best results.

| Dataset | Model | BLEU-1 | BLEU-2 | BLEU-3 | BLEU-4 | CIDEr | ROUGE-L |
|---|---|---|---|---|---|---|---|
| IU X-ray | Base | 0.372 | 0.239 | 0.172 | 0.129 | 0.365 | 0.348 |
| | Base+DKE | 0.440 | 0.286 | 0.206 | 0.156 | 0.378 | 0.378 |
| | Base+SKE | 0.517 | 0.331 | 0.244 | 0.187 | 0.397 | 0.396 |
| | Ours | **0.520** | **0.357** | **0.264** | **0.201** | **0.430** | **0.414** |
| MIMIC-CXR | Base | | | | | | |
| | Base+DKE | | | | | | |
| | Base+SKE | | | | | | |
| | Ours | | | | | | |

Tab. 8 We performed ablation experiments on our proposed method using the IU X-ray and MIMIC-CXR datasets. The base model represents our method implemented on the standard transformer model. "+" represents the combination of proposed modules. Bold represents the best results. AS[35] and KG[2] represent fixed auxiliary signals and static knowledge graphs respectively. Bold represents the best results.

| Dataset | Model | BLEU-1 | BLEU-2 | BLEU-3 | BLEU-4 | CIDEr | ROUGE-L |
|---|---|---|---|---|---|---|---|
| IU X-ray | Base | 0.372 | 0.239 | 0.172 | 0.129 | 0.365 | 0.348 |
| | Base+DKE | 0.440 | 0.286 | 0.206 | 0.156 | 0.378 | 0.378 |
| | Base+AS | 0.388 | 0.264 | 0.153 | 0.108 | 0.334 | 0.357 |
| | Base+SKE | 0.517 | 0.331 | 0.244 | 0.187 | 0.397 | 0.396 |
| | Base+KG | 0.447 | 0.296 | 0.223 | 0.124 | 0.356 | 0.337 |
| | Ours | **0.520** | **0.357** | **0.264** | **0.201** | **0.430** | **0.414** |

|         |         |
|---------|---------|
|         | Base    |
|         | Base+DKE |
|         | Base+AS |
| MIMIC-CXR | Base+SKE |
|         | Base+KG |
|         | Ours    |

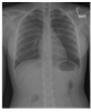

Fig. 6 Performance comparison of radiology diagnostic reports generated by our model and the individual modules. "Base" represents the foundational model. "Base+DKE" signifies the incorporation of disease hashtags. "Base+SKE" denotes the integration of knowledge graphs. "Ours" represents the entirety of the model we proposed.

## 5．Conclusion and Outlook

In this paper, we introduce a model for the automatic generation of medical diagnostic reports, named DMDK. The DMDK model integrates dynamic knowledge from multiple domains to enhance the generation of high-quality reports. Specifically, we design two domains of dynamic knowledge for radiology diagnostic report generation. Dynamic knowledge not only improves the quality of diagnostic reports but also enhances the interpretability of the reports. A large number of experiments prove the superiority of our proposed DMDK model. In the future, we will explore the contribution of different fields of knowledge in the task of generating diagnostic reports to further improve the accuracy of report generation.

# Acknowledgements